\documentclass{article}

\usepackage[table]{xcolor}
\usepackage{iclr2024_conference,times}

\usepackage[noend]{algorithm2e}

\newcommand\blfootnote[1]{%
  \begin{NoHyper}%
  \renewcommand\thefootnote{}\footnote{#1}%
  \addtocounter{footnote}{-1}%
  \end{NoHyper}%
}
\SetCommentSty{mycommfont}
\RestyleAlgo{ruled}
\usepackage[pdftex]{graphicx} 
\usepackage[colorlinks=true, allcolors=blue]{hyperref}     
\usepackage{url}            
\usepackage{wrapfig}
\usepackage{amsmath}
\usepackage{amssymb}
\usepackage{caption} 
\usepackage{makecell}
\usepackage{float}
\usepackage{subcaption}
\usepackage{booktabs}
\usepackage[noabbrev,capitalize]{cleveref}
\usepackage{multirow}
\usepackage[inline,shortlabels]{enumitem}
\usepackage{etoolbox}
\usepackage{array}
\usepackage{tikz}
\usepackage{algcompatible}
\makeatletter
\patchcmd{\@algocf@start}
  {-1.5em}
  {0pt}
  {}{}
\makeatother

\usepackage{siunitx}
\definecolor{rebuttal_color}{rgb}{0.5, 0.0, 0.5}

\author{
Julie Keisler \textsuperscript{* 1,2}, Etienne Le Naour\textsuperscript{* 1,3} \\
\textsuperscript{1} EDF R\&D, Palaiseau, France\\
\textsuperscript{2} INRIA Lille Nord Europe\\
\textsuperscript{3} Sorbonne Université, CNRS, ISIR, 75005 Paris, France\\
\textit{\{julie.keisler, etienne.le-naour\}@edf.fr}
}

\iclrfinalcopy

\makeatletter
\g@addto@macro{\endtabular}{\rowfont{}}
\makeatother
\newcommand{\rowfonttype}{}
\newcommand{\rowfont}[1]{
\gdef\rowfonttype{#1}#1\ignorespaces%
}
\makeatother

\title{WindDragon: Enhancing Wind Power Forecasting with Automated Deep Learning}

\begin{document}

\maketitle
\blfootnote{* Equal contribution}

\begin{abstract}
Achieving net zero carbon emissions by 2050 requires the integration of increasing amounts of wind power into power grids. This energy source poses a challenge to system operators due to its variability and uncertainty. Therefore, accurate forecasting of wind power is critical for grid operation and system balancing. This paper presents an innovative approach to short-term (1 to 6 hour horizon) wind power forecasting at a national level. The method leverages Automated Deep Learning combined with Numerical Weather Predictions wind speed maps to accurately forecast wind power.
\end{abstract}
\section{Introduction}

To meet the 2050 net zero scenario envisaged by the Paris Agreement \citep{paris_agreement_2015}, wind power stands out as a critical energy source for the future. Remarkable progress has been made since 2010, when global electricity generation from wind power was 342 TWh, rising to 2,100 TWh in 2022 \citep{iea_wind_2030}. The IEA targets approximately 7,400 TWh of wind-generated electricity by 2030 to meet the zero-emissions scenario. However, to realize the full potential of this intermittent energy source, accurate forecasts of wind power generation are needed to efficiently integrate it into the power grid. 

Research in wind power forecasting has developed a wide range of methods \citep{giebel2017wind, TAWN2022111758}, including statistical \citep{riahy2008short}, physical \citep{lange2006physical}, hybrid \citep{shi2012evaluation}, and deep learning (DL) \citep{WANG2021117766} approaches. These methods use a variety of data sources, including historical wind power records, geospatial satellite data, on-site camera imagery, and numerical weather prediction (NWP) forecasts. Among these, typical NWP-based methods primarily focus on using local time series of wind speed forecasts for local wind power prediction \citep{piotrowski2022advanced}. However, NWP forecasts produce richer outputs, notably spatial predictions of physical quantities such as wind speed and direction over large scale grids (e.g. national or regional). Predicting aggregated (e.g national or regional) wind power from such fine-grained spatial information appears promising and is largely unexplored in the literature \citep{higashiyama2018feature}. Thus, we propose to explore how wind speed maps combined with suitable machine learning models can capture complex patterns, improving large scale wind power predictions.

\begin{figure}[h!]
    \centering
    \includegraphics[width=0.85\linewidth]{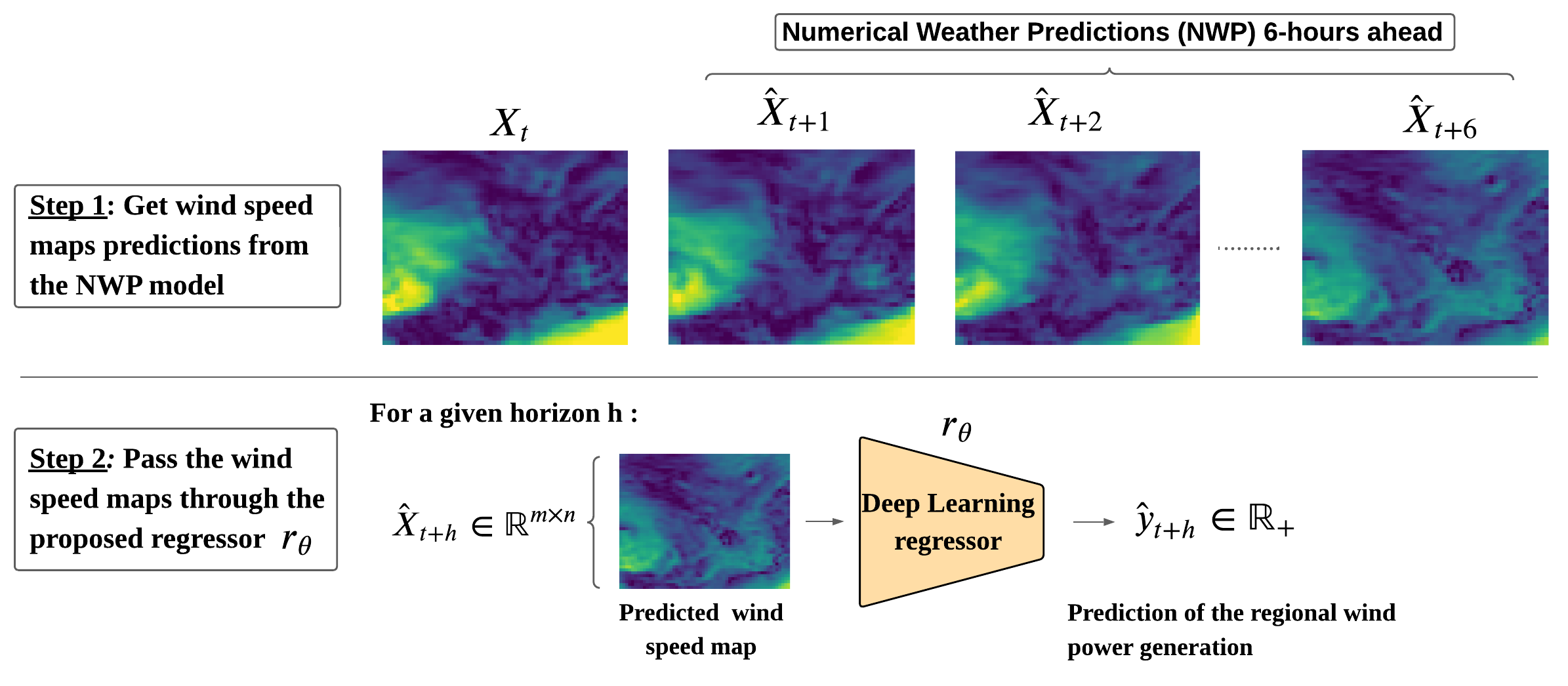}
    \caption{Global scheme for wind power forecasting. Every 6 hours, the NWP model produces hourly forecasts.
    Each map is processed independently by the regressor which maps the grid to the wind power corresponding to the same timestamp.}
    \label{fig:wind_power_forecast_global_procedure}
\end{figure}

In this work, we propose to leverage the spatial information in NWP wind speed maps for national wind power forecasting by exploiting the capabilities of DL models. The overall methodology is illustrated in \cref{fig:wind_power_forecast_global_procedure}. To fully exploit DL mechanisms potential, we introduce WindDragon, an adaptation of the DRAGON\footnote{\url{https://dragon-tutorial.readthedocs.io/en/latest/index.html}} \citep{keisler2023algorithmic} framework. WindDragon is an Automated Deep Learning (AutoDL) framework for short-term wind power forecasting using NWP wind speed maps. WindDragon's performances are benchmarked against conventional computer vision models, such as Convolutional Neural Networks (CNNs) and Vision Transformers (ViTs), as well as standard baselines in wind power forecasting. The experimental results highlight two findings:

\begin{itemize} 
\item The use of full NWP wind speed maps coupled with DL regressors significantly outperforms other baselines. 
\item WindDragon demonstrates superior performance compared to traditional computer vision DL models.
 \end{itemize}

\section{WindDragon: a framework for regression on wind speed maps}
\label{sec:method}

Deep Learning models have the ability to capture complex spatial patterns, which makes them well suited for modeling non-linear relationships between meteorological features and wind energy production. These models are especially useful when wind farms are scattered across the map (see \cref{fig:hull}) and wind speed has significant variance across locations.

\usetikzlibrary{arrows,shapes,positioning,shadows,trees, 3d, calc, shapes.misc, patterns}
\tikzset{
    in_out/.style = {draw, thick, rectangle, minimum height = 0.6cm,  minimum width = 1.2cm, rounded corners=5, fill=gray!30, align=center},
    archi_layer/.style = {draw, thick, circle, align=center},
    a_dia/.style = {draw, thick, diamond, align=center},
    input/.style = {coordinate},
    num_circle/.style = {draw, thick, circle, black}}

\definecolor{input_purple}{HTML}{431cce} 
\definecolor{middle_beige}{HTML}{FAE7D6}
\definecolor{output_red}{HTML}{CE1C4E}
\definecolor{text_white}{HTML}{ECECEC}

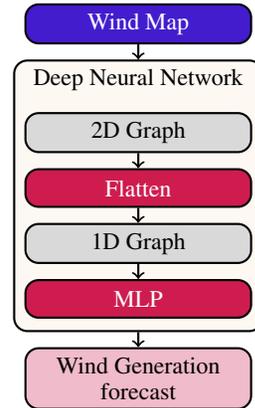
\begin{wrapfigure}[19]{R}{0.28\linewidth}
    \centering
    \begin{tikzpicture}[auto, thick, font=\footnotesize]
        \draw
        node [in_out, minimum height = 0.5cm, minimum width=3cm, fill=input_purple, text=white](input) {Wind Map}
        node [in_out, minimum height = 3.6cm, minimum width=3.3cm, below=0.2cm of input, fill=middle_beige!30, label={[align=center, anchor=north]north:Deep Neural Network}] (dnn) {}
        node [in_out, minimum height = 0.5cm, minimum width=3cm , below=0.9cm of input] (dag1) {2D Graph}
         node [in_out, minimum height = 0.5cm, minimum width=3cm, below=0.2cm of dag1, fill=output_red, text=white] (mlp1) {Flatten}
        node [in_out, minimum height = 0.5cm, minimum width=3cm , below=0.2cm of mlp1] (dag2) {1D Graph}
         node [in_out, minimum height = 0.5cm, minimum width=3cm, below=0.2cm of dag2, fill=output_red, text=white] (mlp2) {MLP}
         node [in_out, minimum height = 0.5cm, minimum width=3cm, below=0.2cm of dnn, fill=output_red!30] (out) {Wind Generation\\ forecast};
        \draw[->](input) -- (dnn); 
        \draw[->](dag1) -- (mlp1);
        \draw[->](mlp1) -- (dag2);
        \draw[->](dag2) -- (mlp2);
        \draw[->](dnn) -- (out);
    \end{tikzpicture}
    \caption{\centering WindDragon’s meta model for wind power forecasting}
    \label{fig:metamodel}
\end{wrapfigure} 

CNNs and ViTs, both prominent in computer vision, might under-perform in the context of wind speed map regression for global wind power forecasting. By learning local and spatial patterns, CNNs efficiently map structured inputs to numerical values. However, CNN's shift-invariant property \citep{zhang2019making} can hinder wind power forecasting because identical wind speeds at different map locations do not equate to the same power generation due to the uneven distribution of wind farms. Conversely, ViTs excel at image classification by segmenting images into patches and applying self-attention mechanisms, but the size of the considered datasets (less than 20000 points for the training dataset) might limit their effectiveness. Given these concerns, the use of AutoDL frameworks to automatically identify the most appropriate DL architecture is a promising solution.

\paragraph{The DRAGON framework.}DRAGON \citep{keisler2023algorithmic} is an AutoDL framework which automatically generates well-performing deep learning models for a given task. Compare to other AutoDL frameworks \citep{liu2019darts, hutter2019automated, autopytorchtab, deng2022efficient}, DRAGON provides a flexible search space, which can be used on any task. It allows the extension of the possibilities in terms of architectures and is adapted when the type of architecture to use is unclear or when high performance is sought by tuning hyperparameters. We used several tools from the generic framework to adapt it for wind power forecasting from wind speed maps.

\paragraph{WindDragon: adapting the DRAGON framework for wind power forecasting.} The neural networks in DRAGON are represented as directed acyclic graphs, with nodes representing the layers and edges representing the connections between them.  In our case, a value $\hat{y_t} \in \mathbb{R}$ is predicted from a 2D map $X_t \in \mathbb{R}^{m \times n}$. The search space is then restricted to a specific family of constrained architectures, as represented in \cref{fig:metamodel}. A first graph processes 2D data and can be composed by convolutions, pooling, normalization, dropout, and attention layers. Then, a flatten layer and a second graph follow. This one is composed by MLPs, self-attention, convolutions and pooling layers. A final MLP layer is added at the end of the model to convert the latent vector to the desired output format. We optimized the solutions from our search space using an evolutionary algorithm, as detailed \cref{section_WindDragon}.

\section{Experiments}
\label{expe_section}

\paragraph{Datasets.}{The wind speed maps used are 100-meter high forecasts at a 9 km resolution provided by the HRES \footnote{\url{https://www.ecmwf.int/en/forecasts/datasets/set-i}} model from the European Centre for Medium-Range Weather Forecasts (ECMWF). The maps are provided at an hourly time step and there are 4 forecast runs per day (every 6 hours). Only the six more recent forecasts are used here as the forecasting horizon of interest is six hours. The hourly french regional and national wind power generation data came from the french TSO\footnote{\url{https://www.rte-france.com/eco2mix}}.}

\paragraph{Data preparation.}{The national forecast of wind power generation is obtained by summing the forecasts of the 12 administrative regions of Metropolitan France. According to our first experiments, this bottom-up technique produced better results than predicting national production directly. 
The division of a national map into regions is a challenge, as shown in \cref{fig:hull} as wind turbines are not evenly distributed across the regions. Therefore, we selected areas around each wind farm in the region and took the convex hull of all the considered points. The result is a seamless map that includes local wind turbines with no gaps to disrupt the models. Installed capacity data for each region - corresponding to the maximum wind power a region can produce - is available and updated every three months. It was collected and used to scale the wind power target. Years from 2018 to 2019 are used to train the models, and data from 2020 is used to evaluate how the models perform.}

\tikzset{every picture/.style={line width=0.75pt}} 

\begin{figure}[h!]
    \centering
    \begin{tikzpicture}[x=0.75pt,y=0.75pt,yscale=-1,xscale=1]
        \draw (330.33,57) node  {\includegraphics[width=120pt,height=90pt]{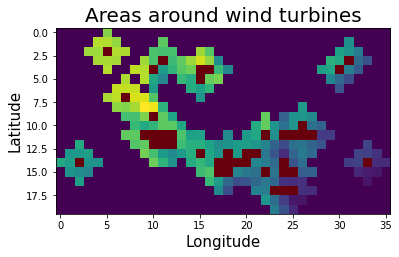}};
        \draw (152.67,57) node  {\includegraphics[width=120pt,height=90pt]{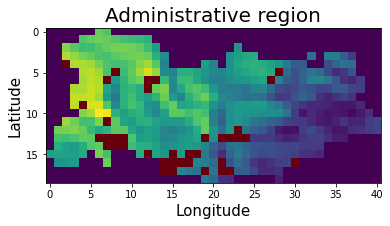}};
        \draw (508.17,57) node  {\includegraphics[width=120pt,height=90pt]{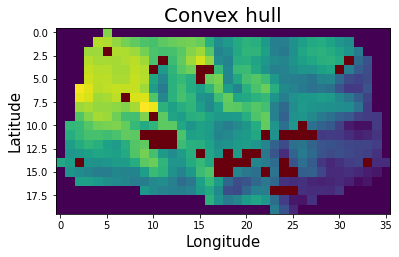}};
        \draw    (229.75,57) -- (250.42,57) ;
        \draw [shift={(252.42,57)}, rotate = 180] [color={rgb, 255:red, 0; green, 0; blue, 0 }  ][line width=0.75]    (9.84,-2.96) .. controls (6.25,-1.25) and (2.97,-0.27) .. (0,0) .. controls (2.97,0.27) and (6.25,1.26) .. (9.84,2.96)   ;
        \draw    (408.25,57) -- (428.92,57) ;
        \draw [shift={(430.92,57)}, rotate = 180] [color={rgb, 255:red, 0; green, 0; blue, 0 }  ][line width=0.75]    (9.84,-2.96) .. controls (6.25,-1.25) and (2.97,-0.27) .. (0,0) .. controls (2.97,0.27) and (6.25,1.26) .. (9.84,2.96)   ;
\end{tikzpicture}
    \caption{Data preparation for the region Auvergne-Rhône-Alpes. The wind farms are represented in red. The first image shows the distribution of wind farms across the administrative region.}
    \label{fig:hull}
\end{figure}
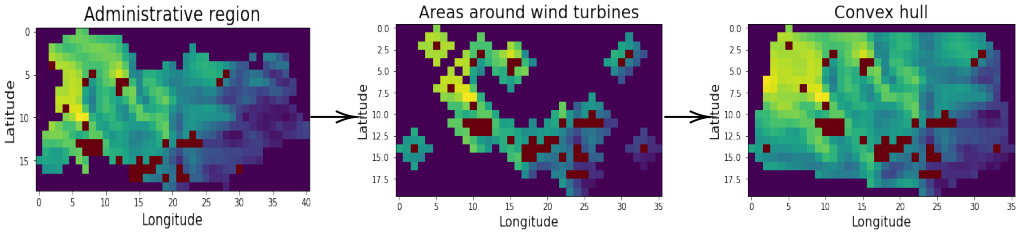

We use the following baselines to compare hourly forecasts for an horizon \( h\) (\( h \in \{1,...,6\} \)):

\begin{itemize}
    \item \textbf{Persistence}: Predicts wind power generation at future time \( t+h \) as equal to the observed generation at current time \( t \). 
    
    \item \textbf{XGB on Wind Speed Mean}: Forecasts wind power at \( t+h \)  using a two-step approach:
    \begin{enumerate*}[(i)]
        \item Compute the mean wind speed for the considered region at \( t+h \) using NWP forecasts.
        \item Apply an XGBoost regressor \citep{chen2016xgboost} to predict power generation based on the computed mean wind speed.
    \end{enumerate*}
    
    \item \textbf{Convolutional Neural Networks (CNNs)}. Forecasts wind power at \( t+h \) using the NWP predicted wind speed map. CNNs can efficiently regress a structured map on a numerical value by learning local and spatial patterns \citep{lecun1995convolutional}.

    \item \textbf{Vision Transformers (ViTs)} Forecasts wind power at \( t+h \) using the NWP predicted wind speed map. The map is segmented into patches and a self-attention mechanism is used to capture the dependencies between these patches \citep{dosovitskiy2020image}.
    
\end{itemize}

The implementation details of the baselines are described in \cref{section_baselines}.

We compute two scores: \textbf{Mean Absolute Error (MAE)} in Megawatts (MW), showing the absolute difference between ground truth and forecast, and \textbf{Normalized Mean Absolute Error (NMAE)}, a percentage obtained by dividing the MAE by the average wind power generation for the test year.

\begin{figure}[h!]
    \centering
    \includegraphics[width=0.99\linewidth]{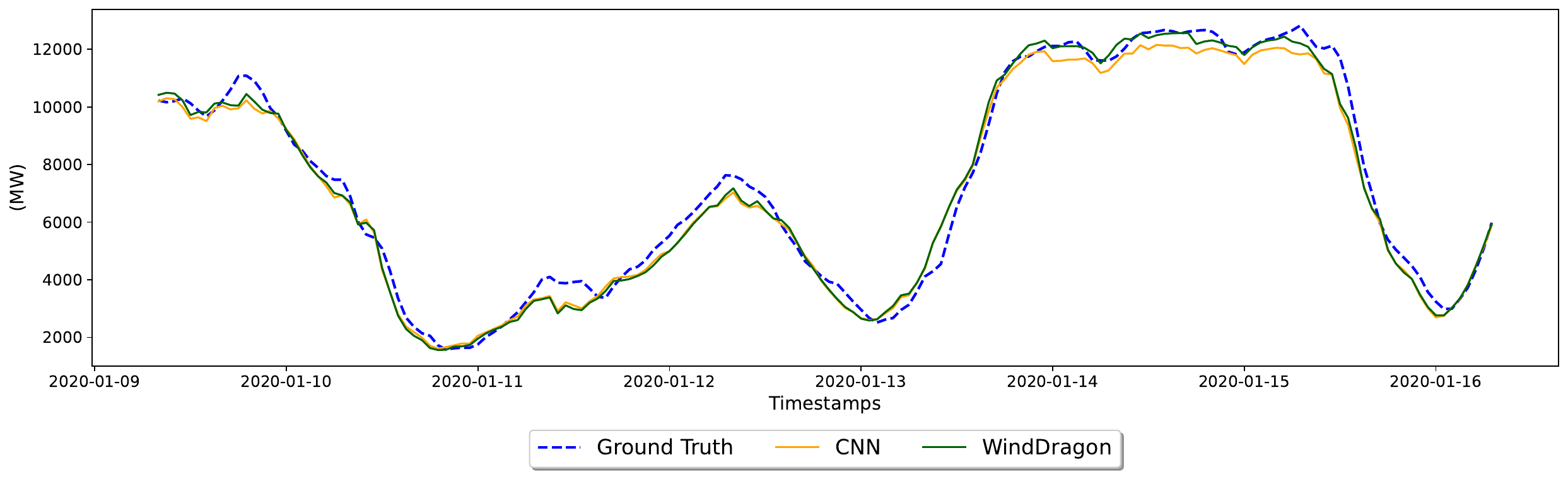}
    \caption{Wind power forecasts for a week in January 2020. The figure displays the ground truth as dotted lines, and the forecasts from the two top-performing models, WindDragon~and~the~CNN.}
    \label{fig:forecast_autoDL}
\end{figure}

\paragraph{Results.}{We run experiments for each of the 12 French metropolitan regions and then aggregate the predictions to derive national results. The national prediction results are presented in \cref{tab:national_results}, while detailed regional results can be found in \cref{tab:regional-results} (\cref{sec_experimental_results}). 

\begin{table}[h!]
    \renewcommand*{\arraystretch}{1.2}
    \setlength{\tabcolsep}{5pt}
    \caption{National results: sum of the regional forecasts for each models. The best results are highlighted in bold and the best second  results are underlined.}
    \label{tab:national_results}
    \begin{center}
    \scalebox{0.75}{%
    \begin{tabular}{lcccccccccc}
        \toprule
        & \multicolumn{2}{c}{WindDragon} & \multicolumn{2}{c}{CNN}  &  \multicolumn{2}{c}{ViT}& \multicolumn{2}{c}{XGB on mean} & \multicolumn{2}{c}{Persistence} \\
        \cmidrule(r){2-3} \cmidrule(r){4-5} \cmidrule(r){6-7} \cmidrule(r){8-9} \cmidrule(r){10-11}
         & MAE (MW) & NMAE & MAE (MW) & NMAE & MAE (MW) & NMAE & MAE (MW) & NMAE & MAE (MW) & NMAE\\
        \midrule
        France & \textbf{346.7} & \textbf{7.7 \%} & \underline{369.0} & \underline{8.1 \%} & 385.7 & 8.5 \%  & 416.7 & 9.2 \% & 779.7 & 17.3 \% \\
        \bottomrule
    \end{tabular}}
    \end{center}
\end{table}

The results in \cref{tab:national_results} highlight three key findings:\begin{enumerate}[(i)]
    \item \textbf{Improved performance with aggregated NWP statistics.} Using the average of NWP-predicted wind speed maps coupled with an XGB regressor significantly outperforms the naive persistence baseline.
    \item \textbf{Gains from full NWP map utilization}. More complex patterns can be captured by using the full predicted wind speed map, as opposed to just the average, thereby improving forecast accuracy. In this context, both the ViT and CNN regressors applied to full maps yielded gains of 31 MW (7.4\%) and 47 MW (11.5\%), respectively, over the mean-based XGB. 
    \item \textbf{WindDragon's superior performances}. WindDragon outperforms all baselines, showing an improvement of 22 MW (6\%) over the CNN. On an annual basis, this corresponds to approximately 193 GWh, which is equivalent to the annual consumption of a French town of 32,000 inhabitants \footnote{based on the average European per capita consumption \citep{statista_europe_electricity_2022}}. Refer to \cref{section_WindDragon} for WindDragon's architecture example.
\end{enumerate}

In \cref{fig:forecast_autoDL}, we present the aggregated national wind power forecasts using both WindDragon and the CNN baseline during a given week. While both models deliver highly accurate forecasts, it's important to highlight that DRAGON demonstrates superior accuracy, particularly in predicting high peak values. See \cref{visual_baseline} for visual comparisons of all baselines performances.
}

\section{Conclusion and impact statement}

In this paper, we have presented two key findings that show great promise. First, using NWP wind speed forecasts as a map significantly improves forecast accuracy compared to using only aggregated values. Second, our framework, WindDragon, shows superior performance to all other baseline models. The significant improvement provided by WindDragon is particularly critical in light of the increasing reliance on wind energy, driven by the pursuit of the net-zero scenario.

Future work could adapt our methodology for photovoltaic (PV) systems, applying it to solar radiation maps generated by NWP models. While current deep learning research in PV primarily focuses on short-term nowcasting \citep{le2022deep}, our method holds promise for extending the forecasting horizon, potentially improving the efficiency and reliability of solar power predictions.

\bibliographystyle{abbrvnat}
\bibliography{bibliography}

\clearpage
\appendix

\section{WINDDRAGON}
\label{section_WindDragon}

\paragraph{Search algorithm.}{The DRAGON framework contains operators, namely mutation and crossover, which are commonly used in meta-heuristics such as the evolutionary algorithm and the simulated annealing, to optimize graphs. The mutation operators are used to add, remove, or modify nodes and connections in the graph, as well as to modify the operations and their hyperparameters within the nodes. Crossover involves exchanging parts of two graphs. The mutation and crossover operators were utilised to construct a steady-state (asynchronous) evolutionary algorithm. Compare to the original algorithm, this version enhances efficiency on HPC by producing two offsprings from the population as soon as a free process is available, rather than waiting for the entire population to be evaluated \citep{liu2018hierarchical}. \\

With the division by region, we slightly modified the generic evolutionary algorithm in WindDragon to avoid having to run an optimisation by region, which would be very costly. In this context, a deep neural network $f$ from our search space $\Omega$ is parametrized by its architecture $\alpha$ and its hyperparemeters $\lambda$. Once $\alpha$ and $\lambda$ have been settled, the model is trained on the data to optimize the weights $\theta$.
We assumed that the architecture $\alpha$ and the hyperparameters $\lambda$ would be broadly similar across regions. Therefore, we modified our evolutionary algorithm to process all regions at the same time. We create and evolve $\alpha$ and $\lambda$ independently of the region, and, to optimize the weights $\theta$, we randomly select the region on which the model would be train and evaluate. In order not to penalize models that have been evaluated on regions that are difficult to predict, we use a global loss function, which consists in dividing the loss obtained on the region $\ell_{\mathrm{region}}$ by the loss of our baseline CNN model on that region, $\mathcal{L}_{\mathrm{region}}$. During the optimisation, for each region, we progressively save the best model evaluated on it. The pseudo code for our steady-state evolutionary algorithm can be found \cref{alg:ssea}.

\begin{algorithm}[h!]
   \caption{Steady-state evolutionary algorithm for multi-regions wind power forecasting}
   \label{alg:ssea}
\begin{algorithmic}
\STATE \textbf{Inputs}: 
\STATE \quad $\Omega$ search space
\STATE \quad $[\mathcal{L}_{\mathrm{region}_1}, \ldots, \mathcal{L}_{\mathrm{region}_R}]$ CNN losses for each region 
\STATE \quad $K$ population size
\STATE \quad $T$ number of iteration
\STATE \textbf{Initialization}  
\STATE \quad Sample $K$ untrained models $f^{\alpha_1, \lambda_1},\dots, f^{\alpha_K, \lambda_K}$ from $\Omega$
\STATE \quad \textbf{For} $k =1,2,\ldots, K$ 
\STATE \quad \quad Select a region $r$ to train the model
\STATE \quad \quad Train $f^{\alpha_k, \lambda_k}$ to get the model weights $\theta^{r}_k$ on the region $r$
\STATE \quad \quad Get the loss $\ell^{k}_{r}$ on this region, and set the model loss to $\ell_k = \ell^{k}_{r} / \mathcal{L}_r$
\STATE \quad \quad \textbf{If} $\ell_r^k$ is the best loss so far on $r$, save $f^{\alpha_k, \lambda_k}_{\theta^r_k}$
\STATE \textbf{For} $t=K+1,K+3, K+5, \ldots, T$ 
\STATE \quad Select two parents $f^{\alpha_{k1}, \lambda_{k1}}$ and $f^{\alpha_{k2}, \lambda_{k2}}$ from the population based on their loss $\ell_k$
\STATE \quad  Mutate and evolve $f^{\alpha_{k1}, \lambda_{k1}}$ and $f^{\alpha_{k2}}$ to $f^{\alpha_{K+t}, \lambda_{K+t}}$ and $f^{\alpha_{K+t+1}, \lambda_{K+t+1}}$
\STATE \quad  Select two regions $r_A$ and $r_B$
\STATE \quad \quad Train respectively $f^{\alpha_{K+t}, \lambda_{K+t}}$ and $f^{\alpha_{K+t+1}, \lambda_{K+t+1}}$ on $r_A$ and $r_B$ to optimize $\theta^{r_A}_{K+t}$ and $\theta^{r_B}_{K+t+1}$
\STATE \quad Get the losses $\ell^{K+t}_{rA}$ and $\ell^{K+t+1}_{rB}$, and set the models losses to $\ell_{K+t} = \ell^{K+t}_{rA} / \mathcal{L}_{rA}$ and $\ell_{K+t+1} = \ell^{K+t+1}_{rB} / \mathcal{L}_{rB}$
\STATE \quad \textbf{If} $\ell_{rA}^{K+t}$ or $\ell_{rB}^{K+t+1}$ are respectively the best losses so far on $r_A$ or $r_B$ save $f^{\alpha_{T+h}, \lambda_{T+h}}_{\theta^{r_A}}$ or $f^{\alpha_{T+h+1}, \lambda_{T+h+1}}_{\theta^{r_B}}$ 
\STATE \quad \textbf{If} $\ell_{K+t}$ or $\ell_{K+t+1}$ are lower than the maximum population loss, we replace the worst model with the new one
\STATE \textbf{Output}: 
\STATE  \quad The best saved model by region
\end{algorithmic}
\end{algorithm}

\begin{figure}[h!]
    \centering
    \includegraphics[height=0.60\linewidth]{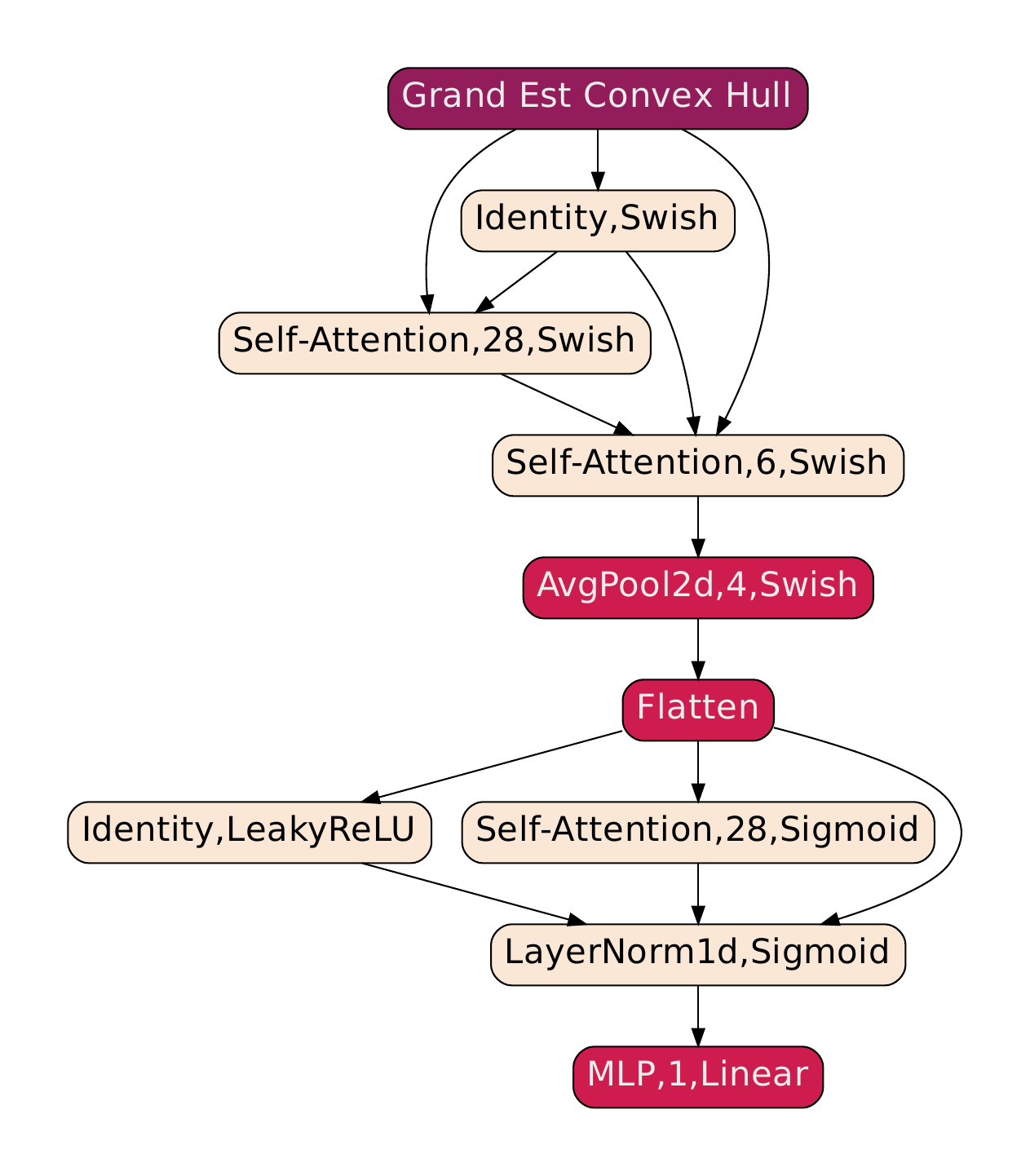}
    \caption{Dragon automatically found architecture applied on the Grand Est region.}
    \label{fig:archi_dragon}
\end{figure}

\paragraph{Results.}{The outputs of WindDragon would be by region the best model found during the optimisation and the prediction of this model. The found architectures vary a bit from a region to another. An example of the best model for the region Grand Est can be found \cref{fig:archi_dragon}. This architecture uses self-attention just like in the Transformer \citep{vaswani2017attention}, but without the patches that can be found in the ViT architecture. The model is also a lot smaller than a Transformer, which can explain why it outperforms the other baselines on this region}

\section{Baselines details}
\label{section_baselines}

The baselines used in \cref{expe_section} are explained in more detail below.

\paragraph{Convolutional Neural Network (CNN).}{\cref{fig:archi_cnn} shows the architecture of the CNN baseline that we implemented. We used a simple grid search to optimize the hyperparameters (e.g. the number of layers, the kernel sizes, the activation functions)}

\begin{figure}[h!]
    \centering
    \includegraphics[height=0.77\linewidth, angle=90]{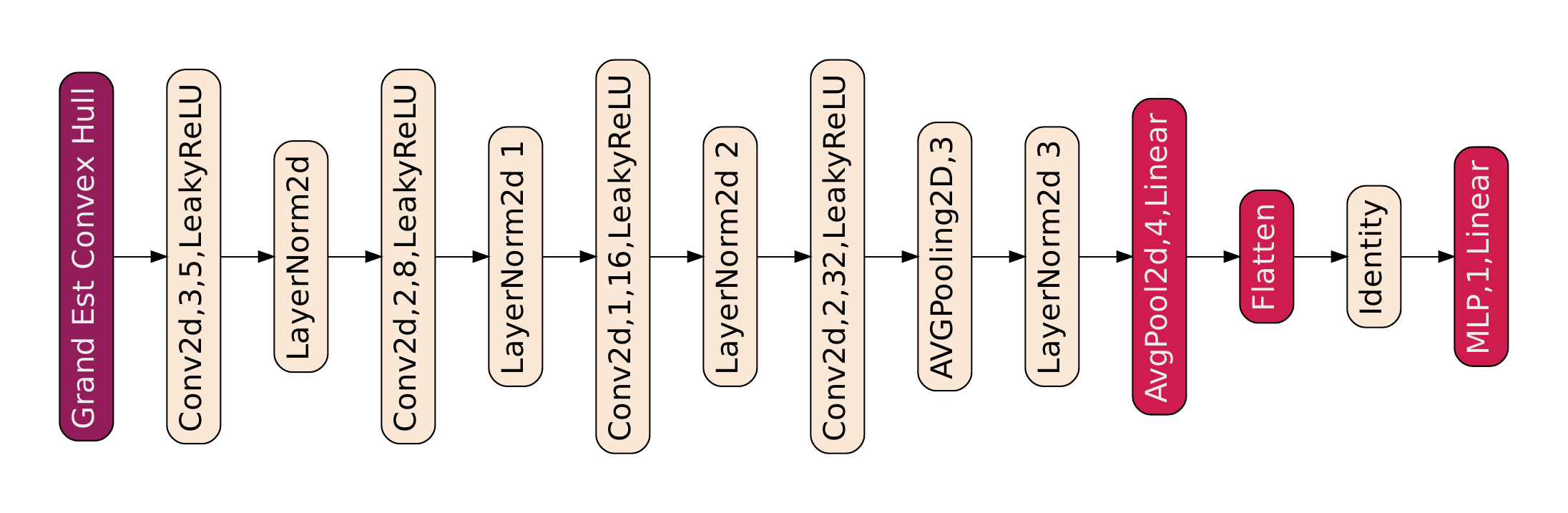}
    \caption{CNN architecture applied on the Grand Est region.}
    \label{fig:archi_cnn}
\end{figure}

\paragraph{Vision Transformer (ViT).}{The Vision Transformer used in this paper is based on SimpleViT's \citep{beyer2022better} architecture. We reused the implementation from lucidrains package \footnote{\url{https://github.com/lucidrains/vit-pytorch}}.} 

\paragraph{XGboost on the mean of the NWP wind speed map.}{\cref{fig:archi_xgb} shows the two-steps procedure of the XGboost baseline.}

\begin{figure}[h!]
    \centering
    \includegraphics[width=0.77\linewidth]{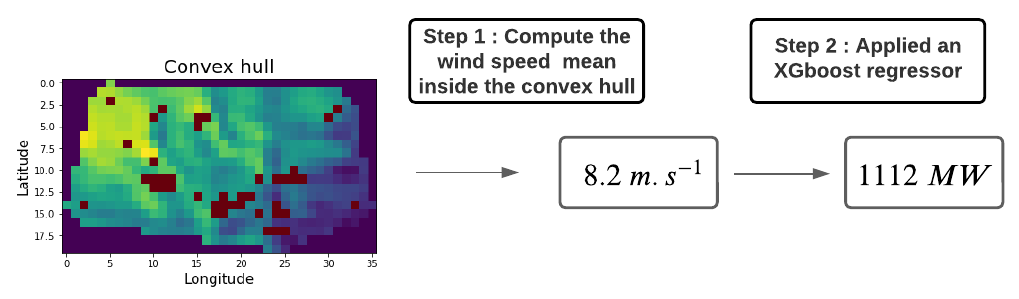}
    \caption{Visual illustration of the XGB two-steps approach on the Auvergne-Rhône-Alpes region.}
    \label{fig:archi_xgb}
\end{figure}

\section{Additional Experimental results}
\label{sec_experimental_results}

\subsection{Regional results}

\begin{table}[H]
    \renewcommand*{\arraystretch}{1.2}
    \setlength{\tabcolsep}{3pt}
    \caption{Regional results. The best results are highlighted in bold and the best second results are underlined.}
    \label{tab:regional-results}
    \begin{center}
    \scalebox{0.63}{%
    \begin{tabular}{lcccccccccc}
        \toprule
        & \multicolumn{2}{c}{WindDragon} & \multicolumn{2}{c}{CNN} & \multicolumn{2}{c}{ViT} & \multicolumn{2}{c}{XGB on mean} & \multicolumn{2}{c}{Persistence} \\
        \cmidrule(r){2-3} \cmidrule(r){4-5} \cmidrule(r){6-7} \cmidrule(r){8-9} \cmidrule(r){10-11}
        Region & {MAE (MW)} & {NMAE} & {MAE (MW)} & {NMAE} & {MAE (MW)} & {NMAE} & {MAE (MW)} & {NMAE} & {MAE (MW)} & {NMAE} \\
        \midrule
        Auvergne-Rhône-Alpes & \textbf{19.5} & \textbf{14.9 \%} & \underline{19.6} & \underline{15.0 \%} & 21.6 & 16.5 \% & 29.2 & 22.4 \% & 28.7 & 22.0 \% \\
        Bourgogne-Franche-Comté & \textbf{32.9} & \textbf{14.8 \%} & \underline{34.1} & \underline{15.4 \%} & 37.2 & 16.8 \% & 42.3 & 19.1 \% & 58.7 & 26.6 \% \\
        Bretagne & \textbf{36.1} & \textbf{14.1} \%& \underline{38.0} & \underline{14.9 \%} & 39.9 & 15.6 \%  & 47.1 & 18.4 \% & 67.2 & 26.3 \% \\
        Centre-Val de Loire & \textbf{53.3} & \textbf{15.0 \%} & \underline{57.3} & \underline{16.1 \%} & 59.0 & 16.6 \%  & 61.9 & 17.5 \% & 96.7 & 27.3 \% \\
        Grand Est & \textbf{125.6} & \textbf{12.5 \%} & \underline{130.5} & \underline{13.1 \%} & 161.0 & 16.1 \% & 148.8 & 14.9 \% & 251.2 & 25.1 \% \\
        Hauts-de-France & \textbf{159.7} & \textbf{12.1 \%} & \underline{167.6} & \underline{12.7 \%} & 177.0 & 13.4 \%  & 178.8 & 13.5 \% & 320.1 & 24.2 \% \\
        Île-de-France & \textbf{6.8} & \textbf{22.6 \%} & \underline{7.17} & \underline{23.7 \%} & 7.4 & 24.3 \%  & 7.5 & 24.9 \% & 9.5 & 31.5 \% \\
        Normandie & \textbf{29.6} & \textbf{12.7 \%} & \underline{30.8} & \underline{13.2 \%} & 31.2 & 13.4 \%  & 36.8 & 15.8 \% & 55.9 & 24.0 \% \\
        Nouvelle-Aquitaine &\textbf{43.1} & \textbf{15.7 \%} & \underline{44.0} & \underline{16.4 \%} & 48.4 & 17.6 \%  & 53.7 & 19.6 \% & 77.9 & 28.4 \% \\
        Occitanie & \textbf{51.2} & \textbf{12.3 \%}& \underline{55.8} & \underline{13.5 \%} & 64.1 & 15.5 \%  & 91.6 & 22.1 \% & 96.3 & 23.2 \% \\
        PACA & \textbf{3.5} & \textbf{32.4 \%} & \textbf{3.5} & \textbf{32.4} \% & 4.0 & 37.2 \%  & 4.5 & 41.4 \% & 4.3 & 39.5 \% \\
        Pays de la Loire & \textbf{37.1} & \textbf{13.6 \%} & \underline{39.0} & \underline{14.3 \%} & 39.9 & 14.7 \%  & 41.9 & 15.4 \% & 74.9 & 27.5 \% \\
        \bottomrule
    \end{tabular}}
    \end{center}
\end{table}

\clearpage

\subsection{Weekly comparative visuals of all baseline results}
\label{visual_baseline}

\begin{figure}[ht]
  \centering
  \begin{subfigure}{\textwidth}
    \centering
    \includegraphics[width=0.82\linewidth]{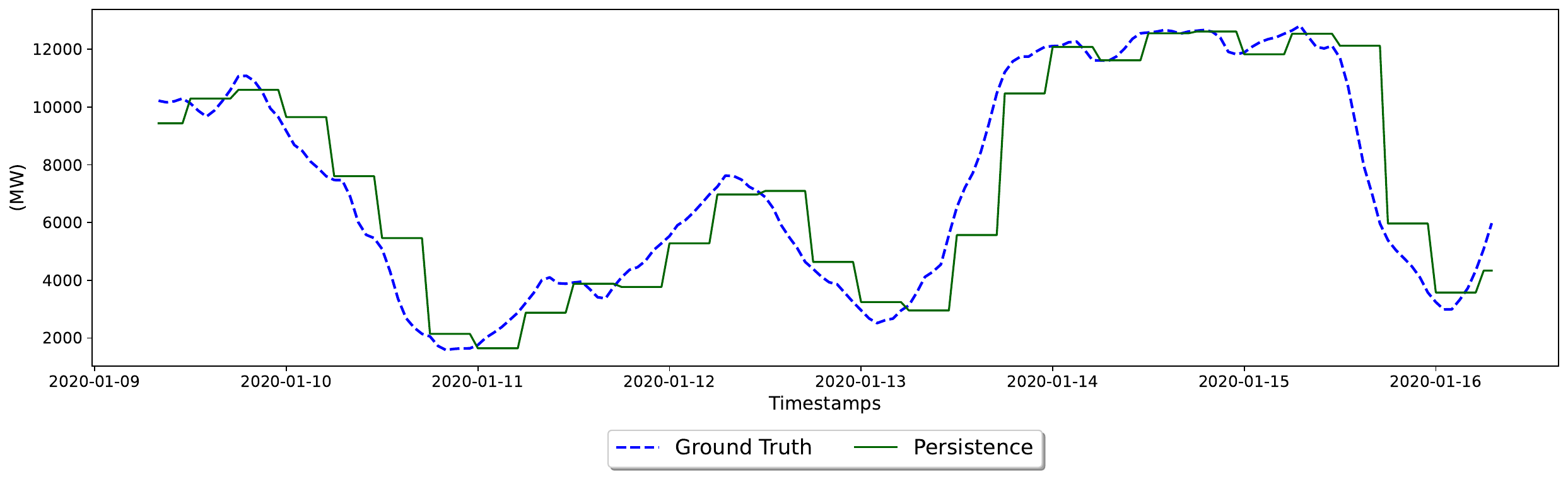}
    \caption{Persistence forecast}
  \end{subfigure}
  \newline
  \begin{subfigure}{\textwidth}
    \centering
    \includegraphics[width=0.82\linewidth]{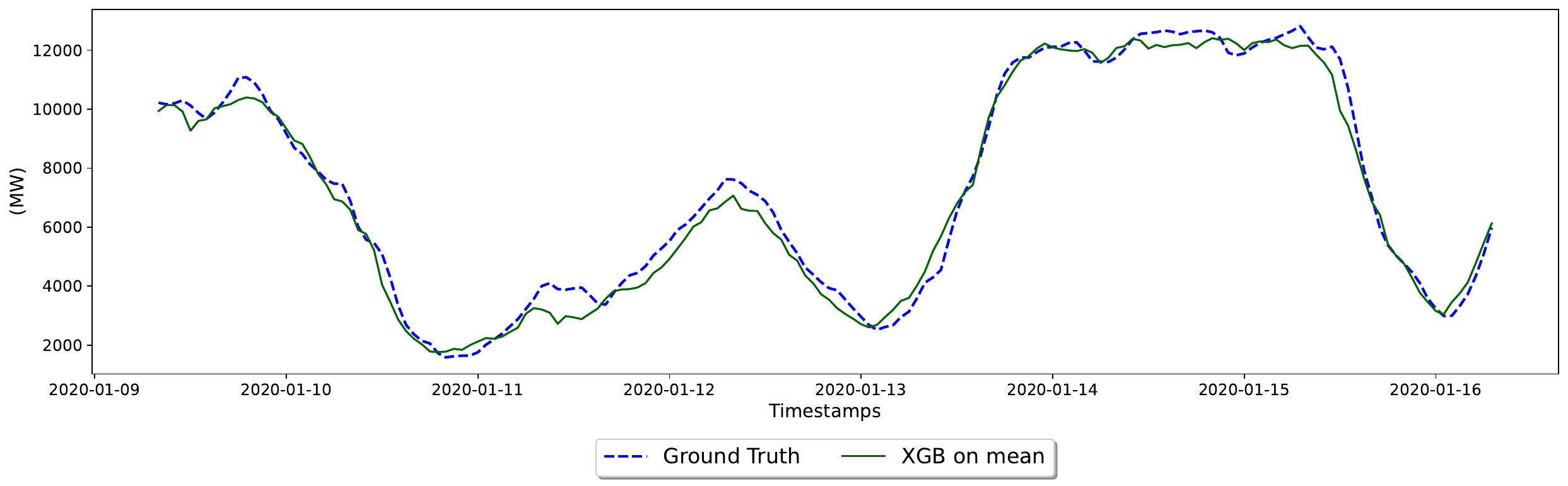}
    \caption{XGB on mean forecast}
  \end{subfigure}
    \newline
  \begin{subfigure}{\textwidth}
    \centering
    \includegraphics[width=0.82\linewidth]{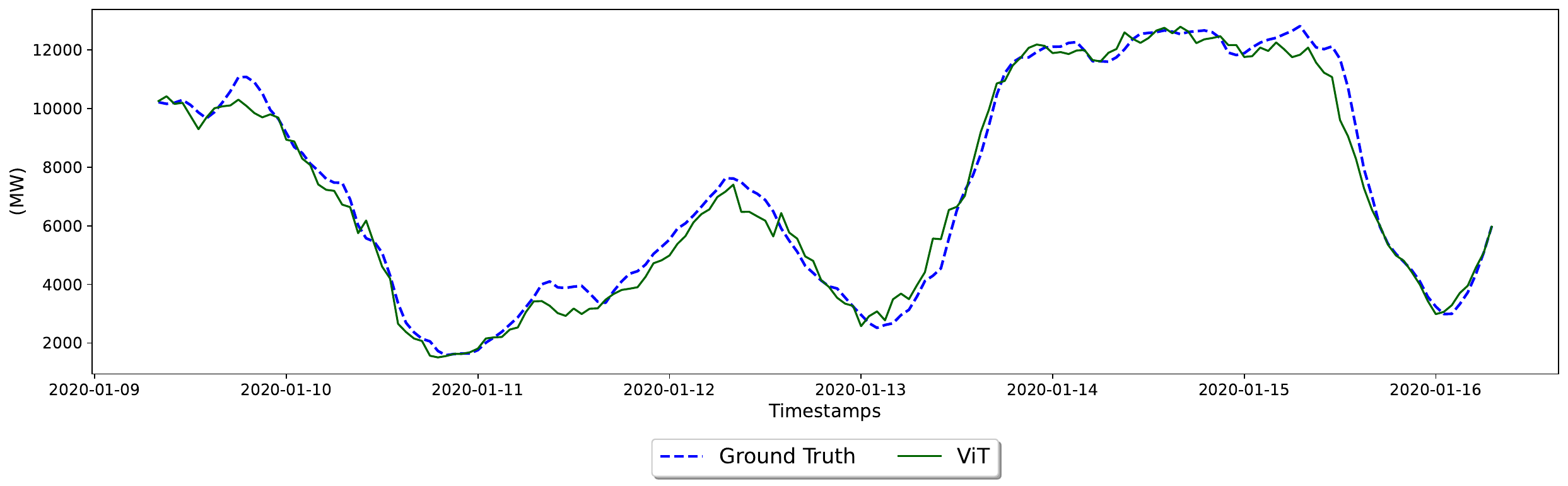}
    \caption{Vision Transformer forecast}
  \end{subfigure}
    \newline
  \begin{subfigure}{\textwidth}
    \centering
    \includegraphics[width=0.82\linewidth]{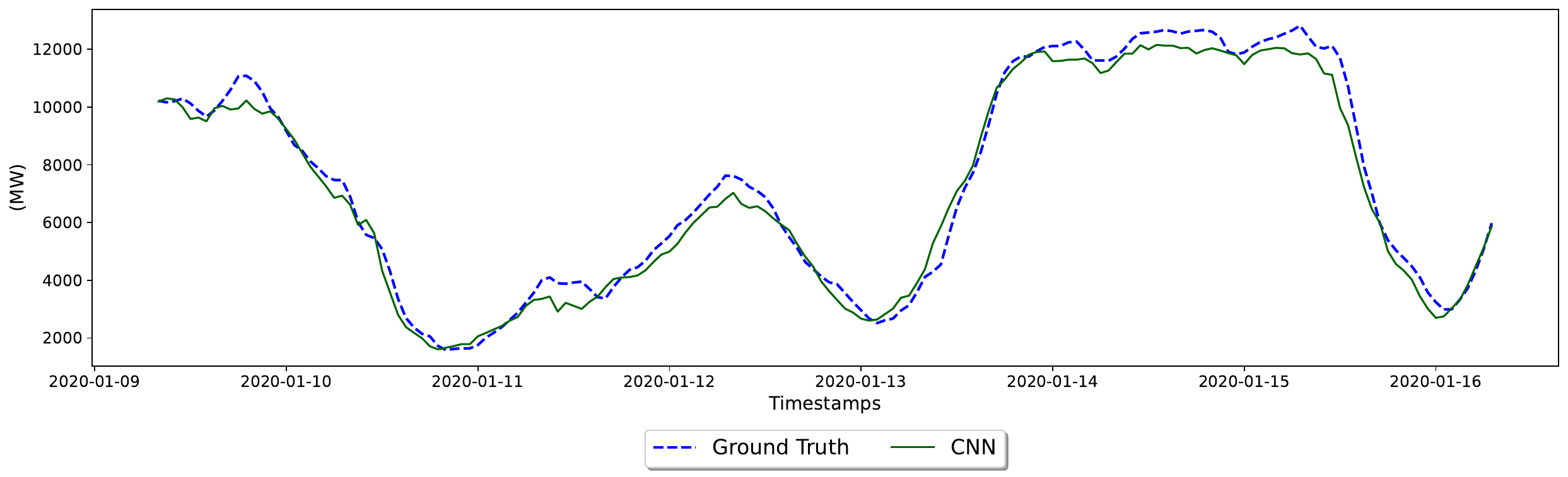}
    \caption{Convolutional Neural Network forecast}
  \end{subfigure}
  \newline
  \begin{subfigure}{\textwidth}
    \centering
    \includegraphics[width=0.82\linewidth]{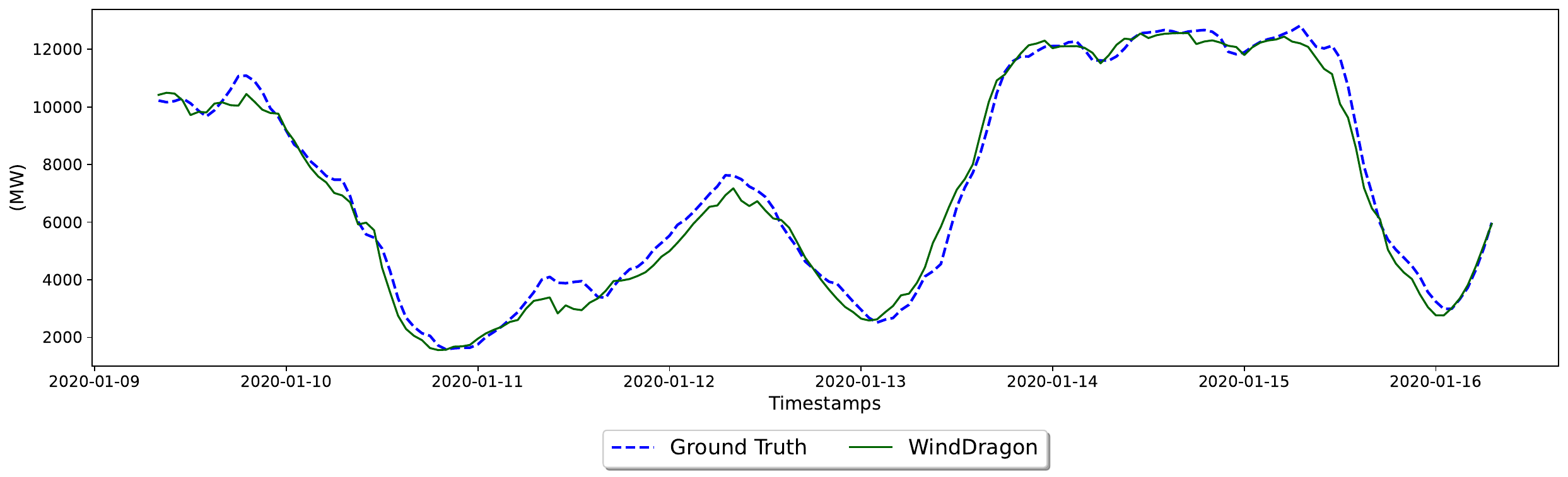}
    \caption{WindDragon forecast}
  \end{subfigure}
  \caption{Weekly comparative visuals}
\end{figure}

\end{document}